# Contrastive Learning Improves Critical Event Prediction in COVID-19 Patients

Tingyi Wanyan[1,2], Hossein Honarvar[1], Suraj K. Jaladanki[1], Chengxi Zang[3], Nidhi Naik[1], Sulaiman Somani[1], Jessica K. De Freitas[1,4], Ishan Paranjpe[1], Akhil Vaid[1], Riccardo Miotto[1], Girish N. Nadkarni[1,5,6], Marinka Zitnik[7], Ariful Azad[2], Fei Wang[3], Ying Ding[8,9], Benjamin S. Glicksberg[1,4*]

1. Hasso Plattner Institute for Digital Health at Mount Sinai, Icahn School of Medicine at Mount Sinai, New York, NY, USA
2. School of Informatics, Computing, and Engineering, Indiana University, Bloomington, IN, USA
3. Department of Population Health Sciences, Weill Cornell Medicine, New York, NY, USA
4. Department of Genetics and Genomic Sciences, Icahn School of Medicine at Mount Sinai, New York, NY, USA
5. Division of Nephrology, Department of Medicine, Icahn School of Medicine at Mount Sinai, New York, NY, USA
6. The Charles Bronfman Institute for Personalized Medicine, Icahn School of Medicine at Mount Sinai, New York, NY
7. Department of Biomedical Informatics, Harvard University, USA
8. Dell Medical School, University of Texas at Austin, Austin, TX, USA
9. School of Informatics, University of Texas at Austin, Austin, TX, USA

## ABSTRACT

### Objective

Machine Learning (ML) models typically require large-scale, balanced training data to be robust, generalizable, and effective in the context of healthcare. This has been a major issue for developing ML models for the coronavirus-disease 2019 (COVID-19) pandemic where data are highly imbalanced. Conventional approaches in ML use cross-entropy loss (CEL) which often suffers from poor margin classification. For the first time, we show that contrastive loss (CL) improves the performance of CEL especially in imbalanced electronic health records (EHR) data for COVID-19 analyses.

### Materials and Methods

We use EHR data to predict mortality, intubation, and intensive care unit (ICU) transfer in hospitalized COVID-19 patients over multiple time windows. We train two sequential architectures (RNN and RETAIN) comparing CEL and CL. Models are tested on the full data set and a restricted data set to simulate higher class imbalance. We further compare clustering ability and feature importance detection between CEL and CL.

### Results

CL models consistently outperform CEL models with the restricted data set on these tasks with differences ranging from 0.04 to 0.15 for AUPRC and 0.05 to 0.1 for AUROC. For the restricted sample, only the CL model maintains proper clustering and is able to identify important features, such as pulse oximetry.

### Discussion and Conclusion

CL outperforms CEL in instances of severe class imbalance, on three EHR outcomes with respect to three performance metrics: predictive power, clustering, and feature importance. We believe that the developed CL framework can be expanded and used for EHR ML work in general.

## INTRODUCTION

### Background and Significance

As of November 2020, coronavirus-disease 2019 (COVID-19) has resulted in over 1.4 million reported deaths with over 250,000 deaths occurring in the United States and over 24,000 deaths in New York City (NYC) [1]. Hospital resources such as medication supply and hospital beds quickly became constrained[2 3]. Due to the novelty of the COVID-19 pandemic, there is a dearth of relevant data available for research purposes, so electronic health records (EHR) became a valuable tool to study and characterize COVID-19 patients. EHR have already been used in biomedical research for disease and patient clustering, patient trajectory modeling, disease prediction, and clinical decision support, among others[4]. Recent studies discovered important findings to better understand COVID-19 through EHR-based analyses[5-11].

Machine learning (ML) is useful to examine resource allocation and risk stratification of patients, and ML models were successfully used to identify high-risk COVID-19 patients[12-21].
For temporal data modeling and prediction of patient outcomes in particular, deep representation learning (DL)[22] holds promise over traditional ML techniques that require manual engineering of informative patient features[23-25]. Heterogeneous graph networks are a powerful DL based graph representation technique that have successfully been utilized in EHR-related studies[26 27]. These models have a graphical structure that captures the underlying relationships between disparate medical concepts such as diagnoses and lab tests[28]. Further, these graph convolutional models can be endowed with an attention mechanism[29] to automatically identify how important local neighbors in the graph are to a given medical concept[30]. Attention models such as RETAIN provide a detailed interpretation of results and maintain high prediction accuracy by using reverse-time attention mechanisms to consolidate past visits[31].

There are, however, substantial concerns about the limited generalizability of these models in COVID-19 (and in general) as they often underperform in external validation[32 33]. Poor generalization of the models is normally due to underspecification[34]. The underlying aspects of EHR data may also limit model effectiveness[35]. Healthcare data sets often have inadequate sample sizes in terms of both small hospitals and rare disease populations and exhibit high class imbalance for key outcomes of interest, such as in rare events, as in the case with COVID-19. Several strategies have been utilized to mitigate these data and modeling challenges including up- and down- sampling, pre-training, transfer learning, and federated learning, but each have their limitations for use in EHR research[36]. Other than these methods, the role of loss function has yet to be thoroughly investigated in the context of COVID-19 EHR work, which is the focus of this work. DL models often use cross-entropy loss (CEL) function, but CEL has been shown to potentially have poor classification margins, making the model less generalizable [37]. Recently, supervised contrastive loss (CL) has been proposed to improve the classification performance of CEL [37]. This original CL algorithm used sets of one anchor, many positives, and many negatives to maximize the similarities within the same class.

Though CL has already been applied to learn visual representations of medical images from paired images and text[38 39], it has not yet been used for EHR data and COVID-19 in particular. Here, we modify a CL-based architecture from its original formulation for EHR tasks. In the original work, CL was used for representation learning and CEL was used for the downstream classification. While we still utilize CL for patient representation learning, we explicitly incorporate classification objective into our CL loss function. This additional term behaves similar to CEL and maximizes the similarities between patient representation (obtained from sequential models) and outcome representation (obtained from heterogeneous relational models)[40].

In this study, we compare different sequential models (RETAIN and RNN) models utilizing CEL and our developed CL in predicting critical outcomes of COVID-19 mortality, ICU transfers, and intubation for patients admitted to a large and diverse patient population from five hospitals within the Mount Sinai Health System in New York. Models are tested on a full data set and a restricted data set with severe class imbalance to better elucidate the impacts of these loss functions on model performance. Results are evaluated within the framework of three dimensions: predictive power, patient clustering, and feature importance.

## MATERIALS AND METHODS

### Clinical Data and Cohort

We obtained Electrical Health Records (EHR) of COVID-19 patients from five hospitals within the Mount Sinai Healthcare System (MSHS). The collected EHR data contain the following information: COVID-19 status, demographics (age, gender, and race), 55 relevant laboratory test results (listed in Supplementary Table 1), vital signs, specifically heart rate, respiration rate, pulse oximetry, blood pressure (diastolic and systolic), temperature, height, and weight, and 12 comorbidities (atrial fibrillation, asthma, coronary artery disease, cancer, chronic kidney disease, chronic obstructive pulmonary disease, diabetes mellitus, heart failure, hypertension, stroke, alcoholism, and liver disease). In addition, we collected information on clinically-relevant outcomes of COVID-19: mortality, discharge, ICU transfer, and intubation. Lab tests and vital signs were measured at multiple time points along the hospital course. In our models, demographics features are used as static features.

### Data Pre-processing

We pre-process the vital signs, lab test, and static features by considering the values between 0.5 and 99.5 percentile to remove any inaccurate measurement. For any numerical data, we normalize the data by calculating the standard score (z-score). For categorical data, we use one-hot encoding representation. Numerical data with missing values are included with zeros.

The initial feature input for vital signs at every time step is the vector $X_v \in R^8$ representing the eight features (Figure 1A). Blood pressure is represented as two features: *systolic* and *diastolic*. Similar to vital signs, the initial feature input for lab tests is the vector $X_l \in R^{55}$ representing the 55 features at each time step. If there is more than one vital sign or lab test for each patient, we average the values of the corresponding time step. We concatenate the vital sign vector and lab test vector to form the vector $X_i \in R^{63}$ as the input at every time step -- the subscripts $v$ and $l$ are dropped for simplicity.

Static features consist of demographics (age, gender, and race) and disease comorbidity information, which are detailed in Supplementary Table 1. For Age, we record the normalized numerical value. We represent the Gender feature as a 2 dimensional (male and female) one-hot vector. We use a 5 dimensional one-hot vector to represent the different groups for Race (African American, White, Asian, Other and Unknown). We represent disease comorbidity as a 12 dimensional one-hot vector. We concatenate all the demographic and disease comorbidity features into the vector $X_d \in R^{20}$.

### Time Sequence Modeling

To model EHR data as a time sequence, we use a previously developed interpretable predictive model named RETAIN[31]. This model is designed specifically to add feature explainability in terms of feature importance score on time sequence data.

For the RETAIN model, we present each patient for $n$ time steps as in Choi et al.[31]:

$$C_p = \sum_{i=1}^{n} \alpha_i \beta_i . v_i \ . \tag{1}$$

where $i$ denotes the time step, $n$ is total number of time steps, $\alpha_i \in R^1$ is the attention vector of weights, $\beta_i \in R^{l \times 1}$ is the attention weight for each feature, $v_i = W_p X_i$ is linear projection of the $m$ dimensional input feature, $X_i \in R^m$ using projection matrix $W_p \in R^{l \times m}$.

As another baseline architecture, we utilize a RNN model's $C_p$ with the original input feature vector $X_i$:

$$C_p = RNN(X_i) \ . \tag{2}$$

**Heterogeneous Relation Modelling**

Instead of treating the outcomes as labels, we model them as a directed heterogeneous bipartite graph as illustrated in Figure 1B. We create a triple relationship between patient, outcome, and event, where the outcome is the relation (or edge) between patient node $p$ and event node $e$. Patient nodes with the same outcome relations are connected to the same event node. Since we are predicting binary outcomes, we have two event nodes representing positive and negative labels. Modeling the data as a bipartite graph provides both label information as well as event and clinical characteristic similarities.

Since the patient and event are two different node types, we use the heterogeneous relational model TransE[40] to project patient and event node types and their outcome relationships into a shared latent space. The TransE model aims to relate different type of nodes by their relationship type and represents the relationship type (outcome) as a translation vector between the two node types. This relation is expressed as:

$$\hat{C}_e = \delta(W_e X_e + b_e)$$
$$C_e = \hat{C}_e - R_o \ , \tag{3}$$

where $X_e$ is the binary outcome representation, $W_e \in R^{l \times 2}$ are learnable projection parameters, and $b_e \in R^l$ are bias parameters, $\delta$ is a non-linear activation function. We use a two dimensional vector $X_e \in (0,1)^2$ to represent positive or negative outcomes. $\hat{C}_e$ is the latent representation of outcomes, $C_e$ is translated representation from $\hat{C}_e$ in the projection space by the learnable translation relational vector $R_o$, which is the relation vectors representing outcome relation that connects patients to positive event and negative event nodes, respectively.
After the projection, we apply similarity comparisons between these two representations ($C_p$ and $C_e$) in the shared latent space.

**Loss Function Delineation**

After the assembly of the bipartite relational graph, we aim to predict the binary outcome of a patient by maximizing the similarity between the binary outcome latent representation and patient representation. The bipartite relational graph also considers the similarities within patient latent representations that connect to the same outcome. Therefore, the objective function is expressed as:

$$\mathcal{L} = \max P\left(N_c(u)|C_p(u)\right) , \tag{4}$$

where $u$ is the patient node of interest in training sample, $C_p(u)$ is the latent representation of patient node $u$. $N_c(u)$ is the patient node's neighboring nodes which consist of binary outcome node representation $C_e$ and similar patient nodes representations $C_p$. In Eq. (4), we optimize the proximity between the representations of center patient node and its neighboring nodes.

The similarity between these latent representations is represented as an inner-product, and we directly apply noise contrastive estimation (NCE) loss to capture the condition probability in Eq. (4)[41].

$$\mathcal{L} = -\sum_{u\in V}\left[\sum_{c_j\in N_c(u)} \log\sigma\left(\vec{c}_j \,.\, \vec{u}\right) + \sum_{j=1}^{K} E_{c_j\sim P_v(c_j)} \log\sigma\left(-\vec{c}_j \,.\, \vec{u}\right)\right], \tag{5}$$

where $V$ is the training patients data set, $\vec{c}_j$ is the latent representation vector of $j$-th context node in $N_c(u)$, $\vec{u}$ is the center node latent representation. $K$ is the number of negative samples, $P_v(c_j)$ is the negative sampling distribution. $\vec{c}_j \,.\, \vec{u}$ are the co-occurrence positive representation pairs, and $\vec{c}_j \,.\, \vec{u}$ are the negative sampling pairs. The non-linear function $\sigma(x) = \frac{1}{1+\exp(-x)}$ captures the similarity score between the representation pairs.

We rewrite the objective function in Eq. (5) in our own notations as follows:

$$\mathcal{L} = -\sum_{u\in V}\left[\sum_{c_{p(j)}\in N_c(u)} \begin{matrix} \log\sigma\left(C_e(u) \,.\, C_p(u)\right) + \log\sigma\left(-C_e^*(u). C_p(u)\right) + \\ \log\sigma\left(C_p(j). C_p(u)\right) + \sum_{j=1}^{K} E_{C_p^*(j)\sim P_v\left(C_p^*(j)\right)} \log\sigma\left(-C_p^*(j). C_p(u)\right) \end{matrix}\right], \tag{6}$$

where $C_e$ is the projected latent representation of the binary outcome node that connects to a given patient representation $C_p$, and the inner product between $C_e$ and $C_p$ measures the similarity between these two representations. Subscript $*$ shows the opposite outcome node of $C_e$ and does not connect to the patient of interest $u$. $C_p(j)$ is the similar context patient node representations that connects to the same outcome node as the patient node $u$. $C_p^*(j)$ are the context patient node representations that connects to the opposite outcome node as the patient node.

The first two terms in Eq. (6) capture the label information between outcome with the patient node, so they function as cross-entropy loss. The last two terms provide additional information of similar patients that connect to the same outcome, and dissimilar patient information that connect to the other outcome.

We set a linear combination rate $a$, and $b$ these two parts to weigh the importance of label information and similar patient information, and our final objective function is as follows:

$$\mathcal{L} = a\left(\mathcal{L}_{ep} + \mathcal{L}_{ep}^*\right) + b(\mathcal{L}_{pp} + \mathcal{L}_{pp}^*)$$

$$\mathcal{L}_{ep} = -\sum\nolimits_{u \in V} \log \sigma\left(C_e(u)\,.\,C_p(u)\right)$$

$$\mathcal{L}_{ep}^* = -\sum\nolimits_{u \in V} \log \sigma\left(-C_e^*(u)\,.\,C_p(u)\right)$$

$$\mathcal{L}_{pp} = -\sum\nolimits_{u \in V} \sum\nolimits_{c_{p(j)} \in N_c(u)} \log \sigma(C_p(j).\,C_p(u))$$

$$\mathcal{L}_{pp}^* = -\sum\nolimits_{u \in V} \sum\nolimits_{j=1}^{K} E_{C_p^*(j) \sim P_v(C_p^*(j))} \log \sigma\left(-C_p^*(j).\,C_p(u)\right)\,. \quad (7)$$

In this work, we set $a = 0.8$, $b = 0.2$ to weigh more towards label information.

After minimizing the CL from Eq. (7), we obtain learned latent representation for events $C_e$, which are used to predict the probability of the events as follows:

$$P\left(Y_e \middle| C_p(u)\right) = \sigma\left(C_e(u)\,.\,C_p(u)\right)\,, \quad (8)$$

where $Y_e$ represents the logit prediction for positive outcomes (mortality, intubation, and ICU transfer).

### Feature Importance Scoring

The linear projection matrix $W_p$ from the RETAIN model allows to interpret variable importance at each time step. Our goal is to predict the probability of the outcome given a center patient representation. We can write this probability the same as Eq. (8).

We can combine Eq. (1), Eq. (3) and Eq. (8) to derive the similarity score as follows:

$$P\left(Y_e \middle| C_p(u)\right) \propto (W_e X_e + b_e - R_o)^T \left(\sum_{i=1}^{n} \alpha_i \beta_i \odot W_p X_i\right) = \sum_{i=1}^{n} \left(\alpha_i \beta_i \odot (X_e^T W_e^T + b_e^T - R_o^T) W_p\right) X_i \ .$$

The contribution score for a specific feature $k$ at time step $i$ for input sample is derived as:

$$\omega\left(Y_e, X_{i,k}\right) = \left(\alpha_i \beta_i \odot (X_e^T W_e^T + b_e^T - R_o^T) W_p\right) X_{i,k}. \tag{9}$$

This is the similarity score between the positive outcome latent representation and patient latent representation for a RETAIN model with CL loss. The larger values of $\omega$ indicates that the feature $k$ has a large contribution towards the prediction result.

For interpretability of a RETAIN model with CEL, we directly compute the importance score in a similar manner as in Choi et al.[31]:

$$\omega\left(y_m, x_{i,k}\right) = \alpha_i W_c \left(\beta_i \odot W_p\right) X_{i,k} \ , \tag{10}$$

where $y_m$ is the label for $m$-th sample.

**Baselines**

In this work, we are evaluating the performance of CL using two time sequence models (RETAIN+CL and RNN+CL). As baseline models, we use the CEL with the same time sequence models (RETAIN+CE and RNN+CE) to evaluate the potential improvements of CL.

As reference and comparison with the CL, the objective function for CEL is as follows:

$$\mathcal{L}_{CEL} = -\frac{1}{N} \sum_{m=1}^{N} \left(y_m \log(\hat{y}_m) + (1 - y_m) \log(1 - \hat{y}_m)\right) \ , \tag{11}$$

where the logit output for the $m$-th sample is:

$$\hat{y}_m = Sigmoid\left(W_c C_p + B_c\right), \tag{12}$$

where $C_p$ is the latent patient representation, $W_e \in R^{2 \times l}$ and $B_c \in R^2$ are the binary projection and bias parameters.

## Experimental Design

We perform three prediction tasks: mortality, ICU transfer, and intubation. We train our models on predicting events for two time frames: 24 and 48 hours before the occurrence of a binary outcome. Longitudinal data (lab tests and vital signs) are binned within windows of 6 hours and averaged if there is more than one measurement per window. The time binning representation is illustrated in Figure 1C. In the training phase, for each positive outcome, we utilize the exact time of an event to generate time frames for the experiments. For patients that did not have these outcome events, we needed a representative frame of reference to align against. Therefore, we compute the mean and standard deviation for the length of time that elapsed from admission for all the affiliated outcomes independently. For patients without an event, we randomly pick a time to use as a reference end point using a Gaussian distribution with the mean and standard deviation obtained from the positive training data. This procedure is shown in Figure 1D.

We also perform the comparisons of outcomes for both full sample and in artificial scenarios of more extreme imbalance (i.e., restricted sample) to determine the extent of performance differences between the two loss functions. In terms of generating the samples for any prediction using CEL, we find that a minimum of 5% positive labels is required to detect both negatives and positives. Therefore, we choose any positive labels percentage to be greater than 5%. For the experiments with full sample, we use all available data where the percentages of positive labels is 23% mortality, 17% ICU transfer, and 10% intubation. For the experiments with restricted sample, we perform down-sampling to reduce the percentage of positive labels by randomly removing a percentage of positive labels. The percentages of positive labels in the down sampled data set are 7% mortality, 7% ICU transfer, and 5% intubation. In the restricted sample, the percentage of each positive label is less than half of the original positive label percentage.

## Training details

In the training stage, after picking a patient node $u$, we select the binary outcome node that connects to $u$. Then we uniformly pick $m$ similar patient nodes that also connect to this outcome node, and we use these samples as positive training pairs. For negative sampling, we first pick the binary outcome node that does not connect to the patient node $u$, and we then uniformly pick $q$ similar patient nodes that connect to this outcome node. We utilize these samples as the negative training pairs. In this work, we use $m = 2$ and $q = 1$ to prioritize the positive samples.

For validation purposes, we perform 10-fold cross validation for the patient of interest and record the mean evaluation values across 10 folds to determine the performance of CL model against the CE model.

All models and data sets are evaluated using the following metrics: area under the receiver-operating characteristic (AUROC), area under the precision, and recall curve (AUPRC). It is important to note that AUPRC is a more reliable metric for imbalanced samples because that takes into account NPV and PPV[42].

## RESULTS

### Model Comparison and Evaluation of Predictive Performance

We first evaluate the results between loss functions using all data (i.e., full sample) for all three tasks, namely mortality, ICU transfer, and intubation prediction. The ROC and PR curves for these cross validated results are shown in Fig.2,3(A-C) and metrics are tabulated in Table 1. For mortality prediction ($23\%$ positive label percentage), we observe that the AUROC and AUPRC scores are similar between CEL and CL. For RETAIN model, under 24h prediction time frame, the AUROC and AUPRC scores are $0.92 \pm 0.01, 0.82 \pm 0.02$ for CEL and $0.92 \pm 0.01, 0.84 \pm 0.02$ for CL. These results indicate that two loss models achieve similar performance when we have relatively more balanced label ratios. For intubation prediction, the positive label percentage is $10.7$ %, which is less than half of the mortality labels. We observe relatively larger performance increases of $0.03 \pm 0.02$ for AUROC score and $0.09 \pm 0.02$ for AUPRC in CL compared to CEL for the RETAIN model. For ICU transfer prediction ($17\%$ positive label percentage) using RETAIN, we observe that the AUROC, AUPRC performance of CL is around $0.84 \pm 0.01, 0.60 \pm 0.02$ which are slightly higher than CEL performance of $0.81 \pm 0.01, 0.57 \pm 0.02$.

To further evaluate the above trends, we conducted an additional experiment to assess how CEL and CL functions perform on the same tasks with more imbalanced outcome ratios. This scenario may be the case in smaller hospital cohorts and for other outcomes. We perform random down-sampling on positive labels (i.e., restricted sample): for mortality, intubation, and ICU transfer prediction tasks, we randomly down-sampled the positive labels to $7\%, 5\%,$ and $7\%$, respectively. The ROC and PR curves are shown in Fig.2,3(E-F) and the performance metrics are recorded in Table 2. The results of the experiment consistently show lower performance using CEL compared to CL. For RETAIN model under 24h time frame, AUROC and AUPRC values are higher for CL than CEL for all outcomes. For instance, the AUROC increases from $0.80 \pm 0.02$ (CEL) to $0.88 \pm 0.02$ (CL) and AUPRC increases from $0.35 \pm 0.03$ (CEL) to $0.45 \pm 0.02$ (CL) for intubation prediction These finding show that under cases with extremely unbalanced label data, models using CL perform better than using CEL.

Our results show that CEL and CL have competitive performance in the full data set but we generally find that for all tasks in restricted data set, models with CEL have lower performance compared to CL (Table 2). Other than the 24-hour time frame, we also perform the same exact analysis for the 48-hour time frame, which shows consistent trends. Finally, we use the RNN as the baseline model other than RETAIN model for all the predictions described earlier and our conclusions hold true (see Table 2).

### Identification of Important Clinical Features

We assess the ability of CEL and CL functions in our models to identify relevant features of interest in full and restricted data sets. Specifically, we performed feature importance score calculations for the RETAIN model on predicting mortality as a representative case.

First, we generated the feature importance scores comparing two models (RETAIN-CL against RETAIN-CEL) over the 24-hour time frame for the full data set for mortality prediction. Fig. 5A and B shows the normalized feature importance heat maps for CL and CEL with the columns representing four 6-hour windows. These heat maps display similar importance scores in terms of key features and their magnitudes. RETAIN identified lab tests and vital sign features that are considered important by both loss models: pulse oximetry (0.26 for CEL, 0.21 for CL); asparate aminotransferase (0.69 for CEL, 0.72 for CL); blood urea nitrogen (0.12 for CEL, 0.12 for CL); lactate (0.24 for CEL, 0.23 for CL); lactate dehydrogenase (0.28 for CEL, 0.3 for CL). All these parameters indicate important aspects of an ill COVID-19 patient.

We then generate feature importance scores using two loss functions for mortality prediction in the restricted data set (i.e., down-sampling the positive label to 7%) for the RETAIN model. We assess how the variable importance score changes under these different conditions for different loss functions. The corresponding heat map is plotted in Fig. 5C and D. We observe that using CL can still capture the highly scored features identified inthe full data set (i.e., weigh similar key features). On the other hand, CEL fails to capture some important features. Of particular interest, the importance of pulse oximetry is no longer prioritized in the restricted sample using CL (importance value is 0.09 for CEL comparing to 0.35 for CL). Also, blood urea nitrogen, lactate have lower importance values of 0.02, 0.09 for CEL compared to 0.15, 0.36 for CL. These findings reaffirm our hypothesis with CL are more robust when the outcome labels are highly imbalanced.

**Visualizing Patient Embeddings**

Finally, we generated 2D t-SNE projections to compare patient embedding representations for RETAIN models between the CL and CEL in predicting all three medical events within 24-hour intervals and the results are shown in Fig. 4. The first two columns show patient embeddings using the full sample data set, and the last two columns show embeddings for the restricted data set. Blue dots represent positive labels and red dots represent negative labels. Models with both loss functions show clear clustering of positive and negative labels in the full data sets. However, when the data set is restricted (i.e., lowered positive labels), the model with CEL consistently show less clear patterns and poorer clustering of patients for all outcomes. In contrast, the RETAIN model with CL maintains its ability to group patients by their clinical outcomes.

## DISCUSSION AND CONCLUSION

In this work, we develop a new DL model based on the CL for predicting three critical events (mortality, intubation, and ICU transfer) for COVID-19 patients using EHR data. Most DL-based EHR analyses utilize CEL as part of modeling. To the best of our knowledge, this is one of the first studies to demonstrate the utility of CL in EHR predictive analyses. We demonstrate the benefit of CL in multiple tasks with imbalanced outcome labels, which is particularly pertinent in the context of COVID-19. We compare the performance of different sequential architectures (RNN and RETAIN) for both CL and the conventional CEL model under two time window horizons. We conduct further experiments for each outcome in a restricted data set of even more imbalanced outcomes and show the benefit of CL is even more pronounced via three separate experimental tasks, namely predictive performance, feature importance, and clustering.

The observed improvements in predictions come from the specific form of CL, which does not only maximize the similarity between patient and outcome embedding representations, but also maximizes the similarities between patient representations related to a specific outcome. However, CEL mainly focuses on maximizing the similarity between patient and outcome one-hot representations. Therefore, CL tends to maximize the margins between classes better than CEL. The better margin classification ability of CL lead to higher performance for imbalanced data, for which the classification is difficult due to poor margin classification of CEL in addition to differences in data distributions for each class.

Our study contains several limitations which need to be addressed. First, we only compare CL to one other loss function, although it is widely used. Second, we only assess two sequential modeling techniques. Another main limitation of our study is excluding several laboratory values due to high levels of missingness, which may impact some of our interpretations. Furthermore, we utilize a specific time sequence modeling representation. Lastly, our work focuses only on one disease use-case. For future studies, other modeling architecture can be assessed to evaluate the generalizability of our CL approach. Also, additional analysis is required to understand feature importance differences between the loss functions and to determine if our CL methodology is applicable to other healthcare data sets. We also plan to assess performance of this strategy for predictive tasks in other diseases. We believe this work represents an effective demonstration of the power of using CL for machine learning work for predictive tasks using EHRs.

Table 1: Binary Outcome Prediction Performance on Different Models using **Full Sample Data** (positive label percentage: 23% for mortality, 10% for intubation, 17% for ICU transfer). CEL: cross-entropy loss; CL: contrastive loss. All predictions are calculated from 10-fold cross validation, for which we record the mean value and standard deviation as confident intervals across folds. Bold values represent best model performance per event.

| Event | Model Type | Time Window | AUROC | AUPRC |
|---|---|---|---|---|
| Mortality | RNN+CE | 24h | 0.91(0.01) | 0.82(0.01) |
| | | 48h | 0.90(0.02) | 0.82(0.02) |
| | RETAIN+CE | 24h | 0.92(0.01) | 0.82(0.02) |
| | | 48h | 0.92(0.01) | 0.83(0.01 |
| | RNN+CL | 24h | 0.91(0.02) | 0.83(0.01) |
| | | 48h | 0.92(0.02) | 0.82(0.01) |
| | RETAIN+CL | 24h | 0.92(0.01) | **0.84**(0.02) |
| | | 48h | **0.93**(0.01) | 0.84(0.01) |
| Intubation | RNN+CE | 24h | 0.83(0.02) | 0.49(0.02) |
| | | 48h | 0.69(0.04) | 0.40(0.03) |
| | RETAIN+CE | 24h | 0.85(0.02) | 0.48(0.02) |
| | | 48h | 0.78(0.03) | 0.39(0.03) |
| | RNN+CL | 24h | **0.88**(0.02) | **0.56**(0.02) |
| | | 48h | 0.84(0.03) | 0.53(0.02) |
| | RETAIN+CL | 24h | 0.88(0.02) | 0.56(0.03) |
| | | 48h | **0.93**(0.01) | 0.51(0.03) |
| ICU Transfer | RNN+CE | 24h | 0.83(0.01) | 0.57(0.02) |
| | | 48h | 0.80(0.01) | 0.54(0.02) |
| | RETAIN+CE | 24h | 0.81(0.02) | 0.57(0.02) |
| | | 48h | 0.81(0.02) | 0.52(0.02) |
| | RNN+CL | 24h | **0.85**(0.01) | **0.62**(0.02) |
| | | 48h | 0.83(0.02) | 0.60(0.02) |
| | RETAIN+CL | 24h | 0.84(0.01) | 0.59(0.02) |
| | | 48h | 0.83(0.01) | 0.59(0.02) |

Table 2: Binary Outcome Prediction Performance on Different Models using **Restricted Sample Data** (positive label percentage: 7% for mortality, 5% for intubation, 7% for ICU transfer). CE: cross-entropy loss; CL: contrastive loss. All predictions are calculated from 10-fold cross validation, for which we record the mean value and standard deviation as confident intervals across folds. Bold values represent best model performance per event.

| Event | Model Type | Time Window | AUROC | AUPRC |
|---|---|---|---|---|
| Mortality | RNN+CE | 24h | 0.83(0.03) | 0.53(0.05) |
| | | 48h | 0.85(0.03) | 0.55(0.04) |
| | RETAIN+CE | 24h | 0.86(0.02) | 0.50(0.05) |
| | | 48h | 0.90(0.03) | 0.53(0.03) |
| | RNN+CL | 24h | 0.89(0.03) | 0.62(0.04) |
| | | 48h | **0.91**(0.03) | 0.63(0.04) |
| | RETAIN+CL | 24h | 0.91(0.01) | 0.59(0.05) |
| | | 48h | 0.91(0.02) | **0.64**(0.04) |
| Intubation | RNN+CE | 24h | 0.79(0.03) | 0.35(0.03) |
| | | 48h | 0.70(0.03) | 0.34(0.04) |
| | RETAIN+CE | 24h | 0.80(0.02) | 0.35(0.03) |
| | | 48h | 0.74(0.02) | 0.31(0.03) |
| | RNN+CL | 24h | **0.88**(0.02) | **0.48**(0.03) |
| | | 48h | 0.83(0.02) | 0.44(0.02) |
| | RETAIN+CL | 24h | 0.88(0.02) | 0.45(0.02) |
| | | 48h | 0.85(0.02) | 0.44(0.03) |
| ICU Transfer | RNN+CE | 24h | 0.78(0.01) | 0.41(0.02) |
| | | 48h | 0.72(0.04) | 0.38(0.03) |
| | RETAIN+CE | 24h | 0.76(0.03) | 0.43(0.04) |
| | | 48h | 0.75(0.04) | 0.43(0.04) |
| | RNN+CL | 24h | 0.81(0.02) | **0.50**(0.03) |
| | | 48h | 0.79(0.01) | 0.47(0.02) |
| | RETAIN+CL | 24h | **0.82**(0.02) | 0.47(0.03) |
| | | 48h | 0.80(0.02) | 0.47(0.03) |

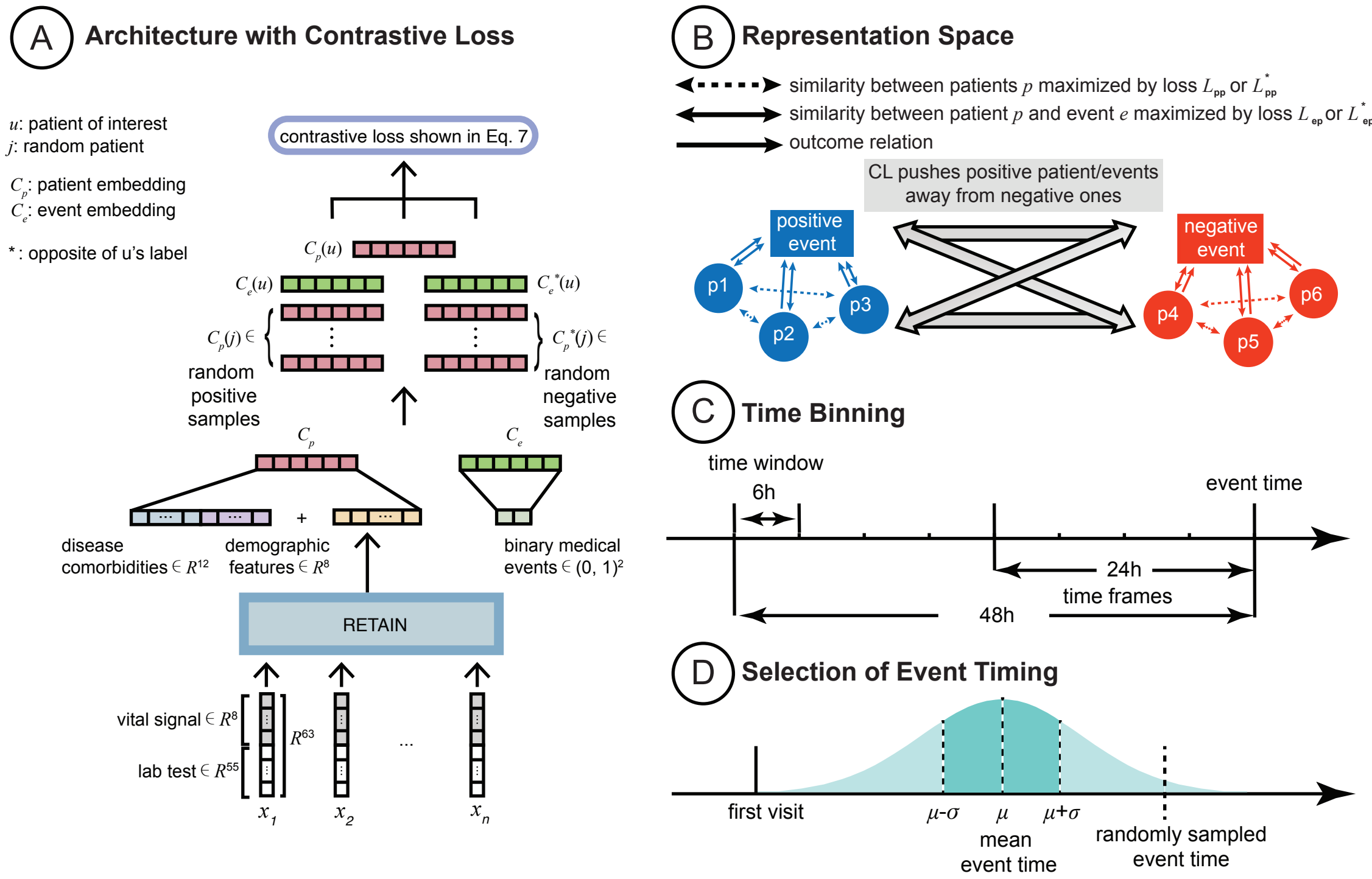

Figure 1: Data and modeling schemas. A) Architecture with Contrastive Loss. Electronic health record (EHR) data are modeled to create patient and event embedding representations which are fed into our contrastive loss (CL) equation. B) Representation Space. CL simultaneously pushes positive patients and event embeddings (i.e., concordant with respect to the outcome of the patient of interest, respectively) away from negative ones. C) Time Binning. Schematic to visualize how we model time sequence. We have two outcome windows (i.e., 24- and 48-hours prior to event) and bin data by 6-hour chunks. D): Selection of Event Timing for Null Outcomes. For patients that do not experience the outcome of interest, we generate a data-driven event time to align against as in C. We compute the mean and standard deviation for the length of time that elapsed from admission for all patients with the affiliated outcomes independently. For patients without an event, we randomly pick a time to use as a reference end point using a Gaussian distribution with the mean and standard deviation obtained from the positive training data.

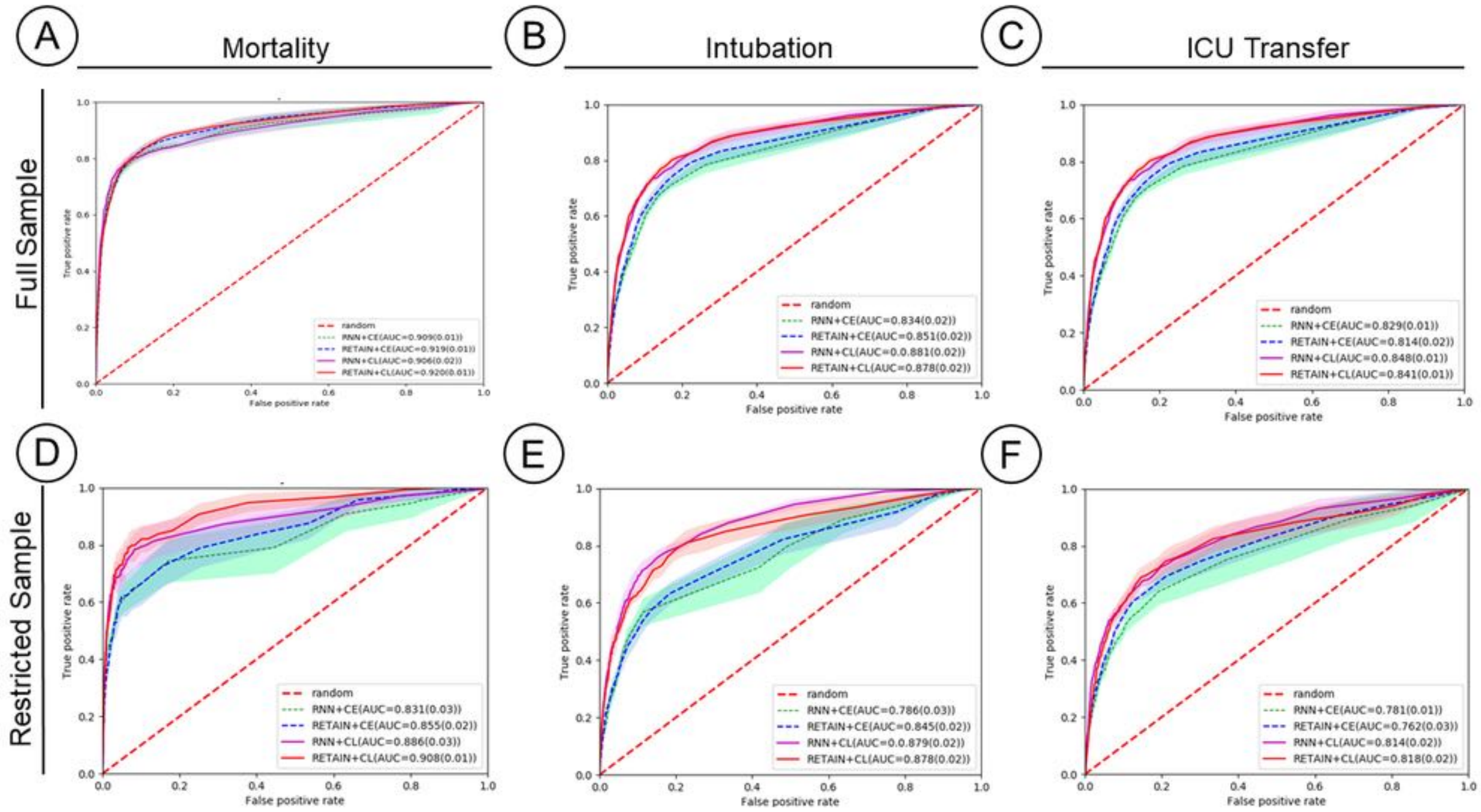

Figure 2: Receiver Operator Curves for All Predictive Tasks in a 24-hour Time Frame. Performance is assessed for both contrastive loss (CL) and cross-entropy loss (CEL) for both RNN and RETAIN modeling strategies. A) Mortality with full data set (23% positive labels); B) Intubation with full data set (11% positive labels); C) ICU transfer with full data set (17% positive labels); D) Mortality with restricted data set (7% positive labels); E) Intubation with restricted data set (5% positive labels); F: ICU transfer with restricted data set (7% positive labels).

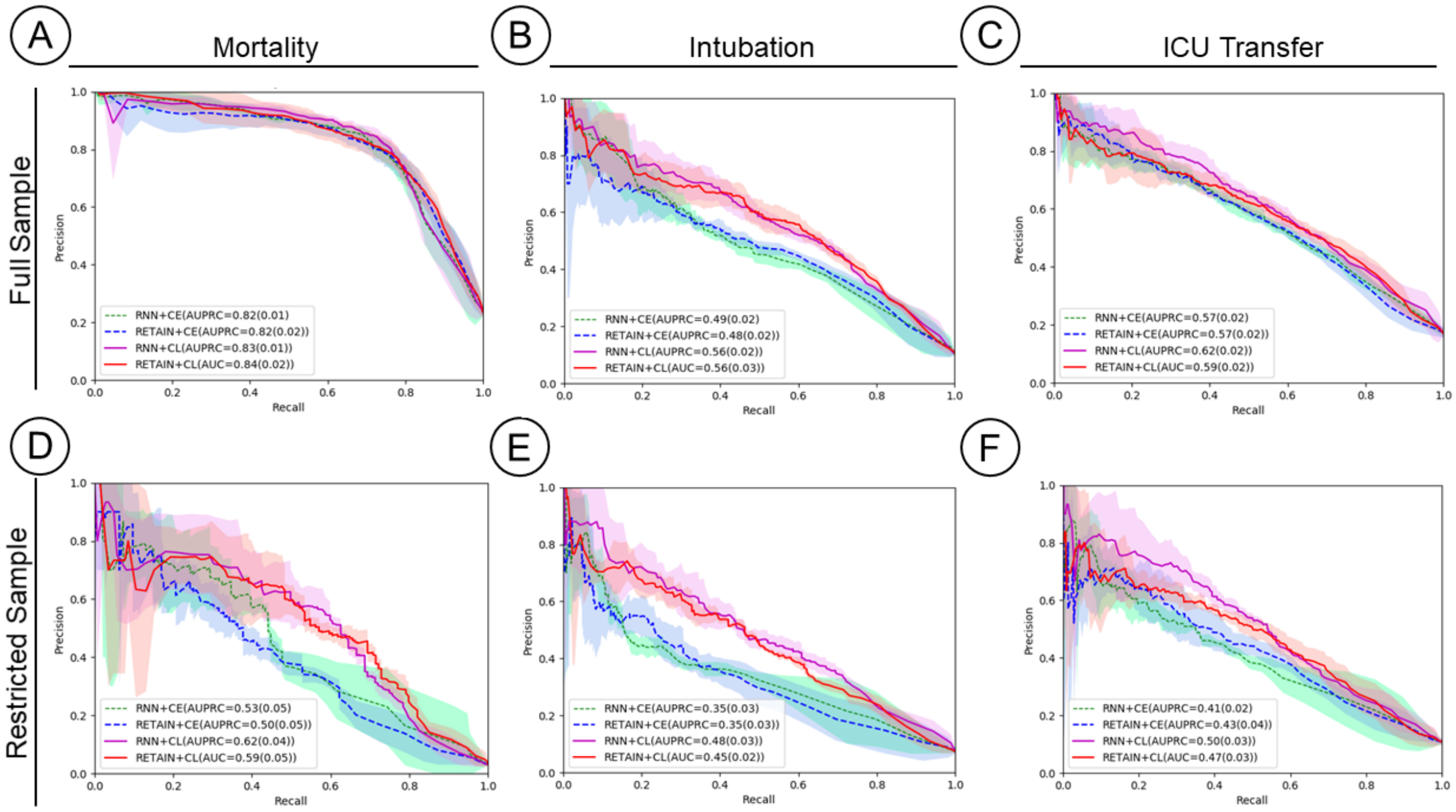

Figure 3: Precision-Recall (PR) Curves for All Event Predictions a 24-hour Time Frame. Performance is assessed for both contrastive loss (CL) and cross-entropy loss (CEL) for both RNN and RETAIN modeling strategies. A) Mortality with full data set (23% positive labels); B) Intubation with full data set (11% positive labels); C):ICU transfer with full data set (17% positive labels); D): Mortality with restricted data set (7% positive labels); E Intubation with restricted data set (5% positive labels); F: ICU transfer with restricted data set (7% positive labels).

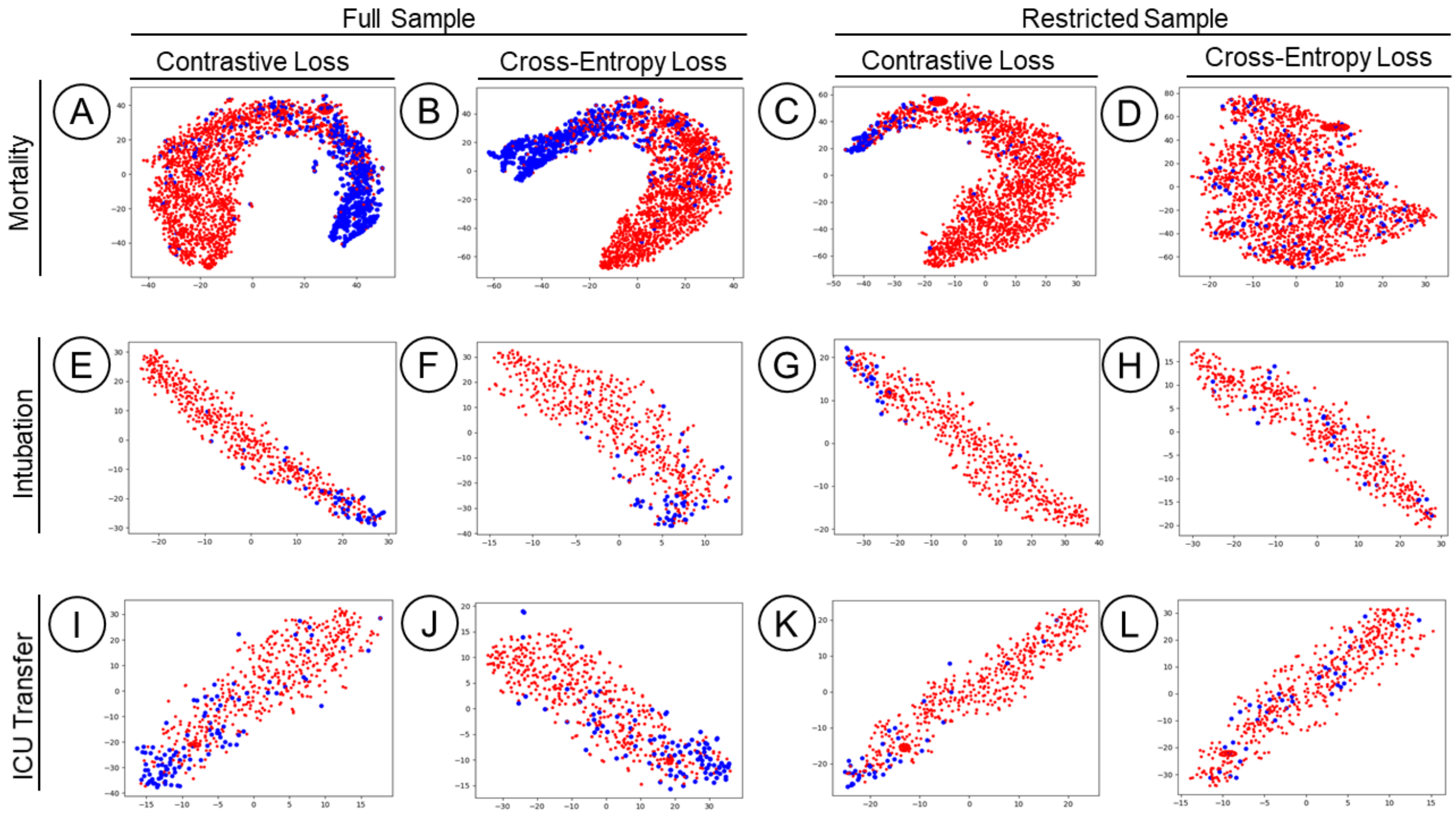

Figure 4: t-SNE Latent Embedding Comparisons for All Event Prediction within 24-hour Time Frame Using RETAIN. Blue dots represent positive labels and red dots represent negative labels. The plot is organized by outcome per row namely 1$^{st}$: mortality, 2$^{nd}$: intubation, and 3$^{rd}$: ICU transfer. The first and third columns represent contrastive loss (CL) plots and the second and forth represent cross-entropy loss (CEL). A) Mortality prediction with CL for the full data set (23% positive labels); B) Mortality prediction with CL for the full data set; C) Mortality prediction with CL for the restricted data set (7% Positive Labels); D) Mortality prediction with CEL for the restricted data set. E) Intubation prediction with CL for the full data set (10% positive labels); F) Intubation prediction with CEL for the full data set; G) Intubation prediction with CL for the restricted data set (5% positive labels); H) Intubation prediction with CEL for the restricted data set. I) ICU transfer prediction with CL for the full data set (17% positive labels); J) ICU transfer prediction with CEL for the full data set; K) ICU transfer prediction with CL for the restricted data set (7% positive labels); L) ICU transfer prediction with CEL for the restricted data set.

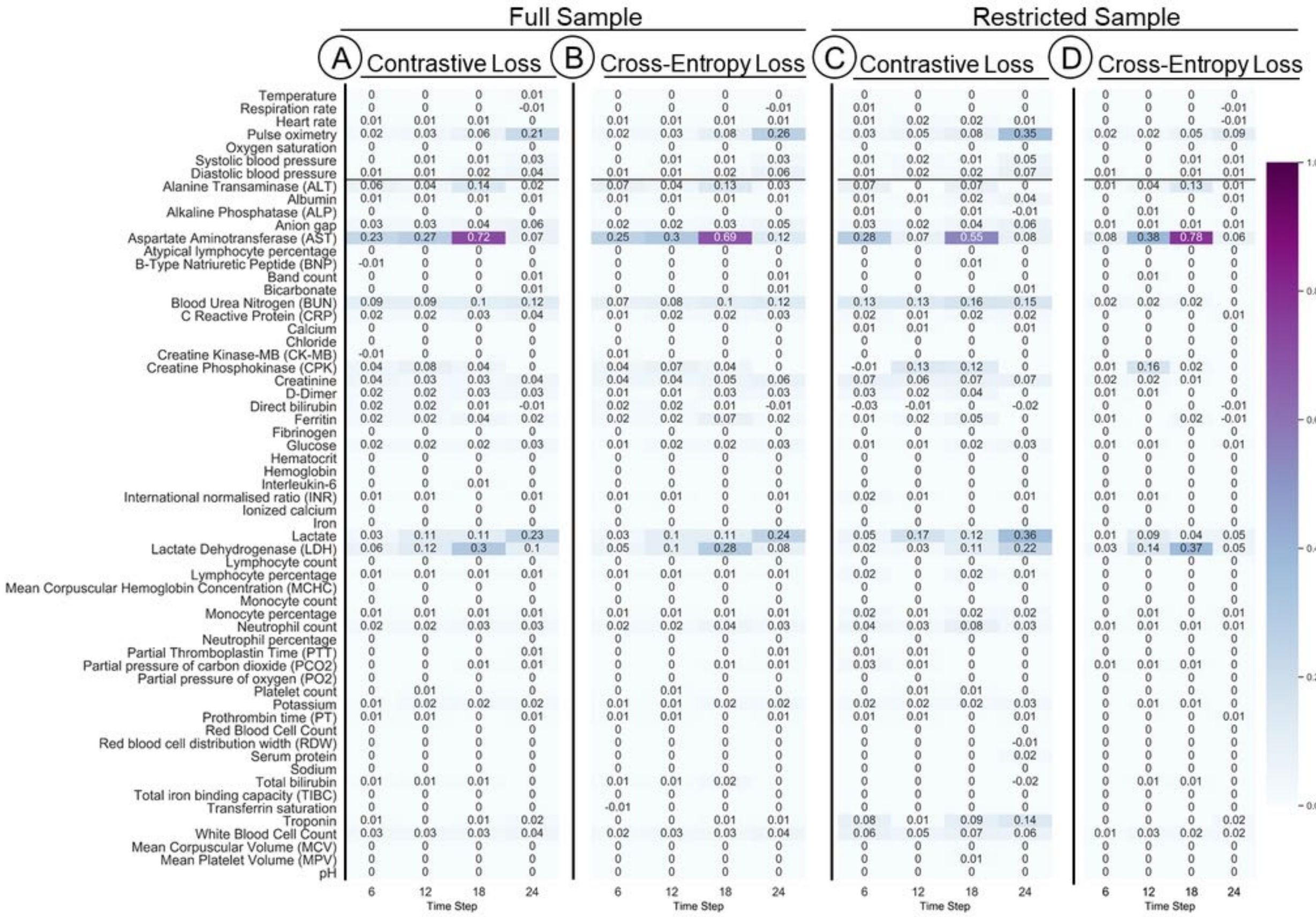

Figure 5: Variable importance in RETAIN mortality prediction at 24 hour intervals utilizing contrastive loss and cross-entropy loss. Each score is the average of 10-fold cross validation. A line separates vitals and laboratory features. A: Full sample (23% positive Labels) in contrastive loss (CL); B: Full Sample (23% positive labels) in cross-entropy loss (CEL); C: Restricted Sample (7% positive Labels) in contrastive loss CL' D: Restricted Sample (7% positive labels) in cross-entropy loss (CEL)

## ACKNOWLEDGEMENTS
We thank the Clinical Data Science and Mount Sinai Data Warehouse teams for providing data. We appreciate all of the providers who contributed to the care of these patients.

## COMPETETING INTERESTS
None.

## FUNDING
This work was supported by U54 TR001433-05, National Center for Advancing Translational Sciences, National Institutes of Health. YD acknowledges support from Amazon Machine Learning Research Award.

## AUTHOR CONTRIBUTIONS
All authors approved the final version of the manuscript.

## REFERENCES

1. Dong E, Du H, Gardner L. An interactive web-based dashboard to track COVID-19 in real time. Lancet Infect Dis 2020;**20**(5):533-34 doi: 10.1016/S1473-3099(20)30120-1[published Online First: Epub Date]|.
2. Thompson CN, Baumgartner J, Pichardo C, et al. COVID-19 Outbreak - New York City, February 29-June 1, 2020. MMWR Morb Mortal Wkly Rep 2020;**69**(46):1725-29 doi: 10.15585/mmwr.mm6946a2[published Online First: Epub Date]|.
3. McMahon DE, Peters GA, Ivers LC, Freeman EE. Global resource shortages during COVID-19: Bad news for low-income countries. PLoS Negl Trop Dis 2020;**14**(7):e0008412 doi: 10.1371/journal.pntd.0008412[published Online First: Epub Date]|.
4. Glicksberg BS, Johnson KW, Dudley JT. The next generation of precision medicine: observational studies, electronic health records, biobanks and continuous monitoring. Hum Mol Genet 2018;**27**(R1):R56-R62 doi: 10.1093/hmg/ddy114[published Online First: Epub Date]|.
5. Clifford CT, Pour TR, Freeman R, et al. Association between COVID-19 diagnosis and presenting chief complaint from New York City triage data. Am J Emerg Med 2020 doi: 10.1016/j.ajem.2020.11.006[published Online First: Epub Date]|.
6. Reeves JJ, Hollandsworth HM, Torriani FJ, et al. Rapid response to COVID-19: health informatics support for outbreak management in an academic health system. J Am Med Inform Assoc 2020;**27**(6):853-59 doi: 10.1093/jamia/ocaa037[published Online First: Epub Date]|.
7. Somani SS, Richter F, Fuster V, et al. Characterization of Patients Who Return to Hospital Following Discharge from Hospitalization for COVID-19. J Gen Intern Med 2020;**35**(10):2838-44 doi: 10.1007/s11606-020-06120-6[published Online First: Epub Date]|.
8. Wagner T, Shweta F, Murugadoss K, et al. Augmented curation of clinical notes from a massive EHR system reveals symptoms of impending COVID-19 diagnosis. Elife 2020;**9** doi: 10.7554/eLife.58227[published Online First: Epub Date]|.
9. Wang Z, Zheutlin A, Kao YH, et al. Hospitalised COVID-19 patients of the Mount Sinai Health System: a retrospective observational study using the electronic medical records. BMJ Open 2020;**10**(10):e040441 doi: 10.1136/bmjopen-2020-040441[published Online First: Epub Date]|.
10. Williamson EJ, Walker AJ, Bhaskaran K, et al. Factors associated with COVID-19-related death using OpenSAFELY. Nature 2020;**584**(7821):430-36 doi: 10.1038/s41586-020-2521-4[published Online First: Epub Date]|.
11. Wu Z, McGoogan JM. Characteristics of and Important Lessons From the Coronavirus Disease 2019 (COVID-19) Outbreak in China: Summary of a Report of 72314 Cases From the Chinese Center for Disease Control and Prevention. JAMA 2020;**323**(13):1239-42 doi: 10.1001/jama.2020.2648[published Online First: Epub Date]|.
12. Yang HS, Hou Y, Vasovic LV, et al. Routine laboratory blood tests predict SARS-CoV-2 infection using machine learning. Clinical chemistry 2020;**66**(11):1396-404
13. Wynants L, Van Calster B, Collins GS, et al. Prediction models for diagnosis and prognosis of covid-19: systematic review and critical appraisal. bmj 2020;**369**

14. Liang W, Liang H, Ou L, et al. Development and validation of a clinical risk score to predict the occurrence of critical illness in hospitalized patients with COVID-19. JAMA Internal Medicine 2020
15. Gysi DM, Valle ÍD, Zitnik M, et al. Network medicine framework for identifying drug repurposing opportunities for covid-19. arXiv preprint arXiv:2004.07229 2020
16. Yan L, Zhang H-T, Goncalves J, et al. An interpretable mortality prediction model for COVID-19 patients. Nature Machine Intelligence 2020:1-6
17. Vaishya R, Javaid M, Khan IH, Haleem A. Artificial Intelligence (AI) applications for COVID-19 pandemic. Diabetes & Metabolic Syndrome: Clinical Research & Reviews 2020
18. Mei X, Lee H-C, Diao K-y, et al. Artificial intelligence–enabled rapid diagnosis of patients with COVID-19. Nature Medicine 2020:1-5
19. Vaid A, Somani S, Russak AJ, et al. Machine Learning to Predict Mortality and Critical Events in a Cohort of Patients With COVID-19 in New York City: Model Development and Validation. Journal of medical Internet research 2020;**22**(11):e24018
20. Alimadadi A, Aryal S, Manandhar I, Munroe PB, Joe B, Cheng X. Artificial intelligence and machine learning to fight COVID-19: American Physiological Society Bethesda, MD, 2020.
21. Zhou Y, Wang F, Tang J, Nussinov R, Cheng F. Artificial intelligence in COVID-19 drug repurposing. The Lancet Digital Health 2020
22. Landi I, Glicksberg BS, Lee HC, et al. Deep representation learning of electronic health records to unlock patient stratification at scale. NPJ Digit Med 2020;**3**:96 doi: 10.1038/s41746-020-0301-z[published Online First: Epub Date]|.
23. Doctor ai: Predicting clinical events via recurrent neural networks. Machine Learning for Healthcare Conference; 2016.
24. Miotto R, Li L, Kidd BA, Dudley JT. Deep patient: an unsupervised representation to predict the future of patients from the electronic health records. Scientific reports 2016;**6**(1):1-10
25. Shickel B, Tighe PJ, Bihorac A, Rashidi P. Deep EHR: a survey of recent advances in deep learning techniques for electronic health record (EHR) analysis. IEEE journal of biomedical and health informatics 2017;**22**(5):1589-604
26. GRAM: graph-based attention model for healthcare representation learning. Proceedings of the 23rd ACM SIGKDD International Conference on Knowledge Discovery and Data Mining; 2017.
27. Learning the Graphical Structure of Electronic Health Records with Graph Convolutional Transformer. Proceedings of the AAAI Conference on Artificial Intelligence; 2020.
28. Heterogeneous Graph Embeddings of Electronic Health Records Improve Critical Care Disease Predictions. International Conference on Artificial Intelligence in Medicine; 2020. Springer.
29. Veličković P, Cucurull G, Casanova A, Romero A, Lio P, Bengio Y. Graph attention networks. arXiv preprint arXiv:1710.10903 2017
30. Heterogeneous graph attention network. The World Wide Web Conference; 2019.
31. Choi E, Bahadori MT, Sun J, Kulas J, Schuetz A, Stewart W. Retain: An interpretable predictive model for healthcare using reverse time attention mechanism. Advances in neural information processing systems 2016;**29**:3504-12
32. Barish M, Bolourani S, Lau LF, Shah S, Zanos TP. External validation demonstrates limited clinical utility of the interpretable mortality prediction model for patients with COVID-19. Nature Machine Intelligence 2020:1-3

33. Gupta RK, Marks M, Samuels TH, et al. Systematic evaluation and external validation of 22 prognostic models among hospitalised adults with COVID-19: An observational cohort study. European Respiratory Journal 2020;**56**(6)
34. D'Amour A, Heller K, Moldovan D, et al. Underspecification presents challenges for credibility in modern machine learning. arXiv preprint arXiv:2011.03395 2020
35. Cyganek B, Graña M, Krawczyk B, et al. A survey of big data issues in electronic health record analysis. Applied Artificial Intelligence 2016;**30**(6):497-520
36. Xu J, Glicksberg BS, Su C, Walker P, Bian J, Wang F. Federated Learning for Healthcare Informatics. J Healthc Inform Res 2020:1-19 doi: 10.1007/s41666-020-00082-4[published Online First: Epub Date]|.
37. Khosla P, Teterwak P, Wang C, et al. Supervised Contrastive Learning. 2020. https://ui.adsabs.harvard.edu/abs/2020arXiv200411362K (accessed April 01, 2020).
38. Li Z, Roberts K, Jiang X, Long Q. Distributed learning from multiple EHR databases: Contextual embedding models for medical events. J Biomed Inform 2019;**92**:103138 doi: 10.1016/j.jbi.2019.103138[published Online First: Epub Date]|.
39. Zhang Y, Jiang H, Miura Y, Manning CD, Langlotz CP. Contrastive Learning of Medical Visual Representations from Paired Images and Text. 2020. https://ui.adsabs.harvard.edu/abs/2020arXiv201000747Z (accessed October 01, 2020).
40. Bordes A, Usunier N, Garcia-Durán A, Weston J, Yakhnenko O. Translating embeddings for modeling multi-relational data. Proceedings of the 26th International Conference on Neural Information Processing Systems - Volume 2. Lake Tahoe, Nevada: Curran Associates Inc., 2013:2787–95.
41. Levy O, Goldberg Y. Neural word embedding as implicit matrix factorization. Advances in neural information processing systems 2014;**27**:2177-85
42. Saito T, Rehmsmeier M. The precision-recall plot is more informative than the ROC plot when evaluating binary classifiers on imbalanced datasets. PloS one 2015;**10**(3):e0118432